\documentclass{article}

\usepackage{microtype}
\usepackage{graphicx}
\usepackage{subcaption}
\usepackage{booktabs} 

\usepackage{stmaryrd}
\usepackage{amsfonts}
\usepackage{latexsym}
\usepackage{amssymb}
\usepackage{bm}
\usepackage{amsmath,empheq}

\usepackage[accepted,nohyperref]{icml2019}

\icmltitlerunning{Exploring Deep Anomaly Detection Methods Based on Capsule Net}

\begin{document}

\twocolumn[
\icmltitle{Exploring Deep Anomaly Detection Methods Based on Capsule Net}




\begin{icmlauthorlist}
\icmlauthor{Xiaoyan Li}{ou}
\icmlauthor{Iluju Kiringa}{ou}
\icmlauthor{Tet Yeap}{ou}
\icmlauthor{Xiaodan Zhu}{qu}
\icmlauthor{Yifeng Li}{nrc}
\end{icmlauthorlist}

\icmlaffiliation{ou}{School of Electrical Engineering and Computer Science, University of Ottawa}
\icmlaffiliation{qu}{Department of Electrical and Computer Engineering, Queen's University}
\icmlaffiliation{nrc}{Digital Technologies Research Centre, National Research Council Canada}
\icmlcorrespondingauthor{Xiaoyan Li}{xli343@uottawa.ca}
\icmlcorrespondingauthor{Yifeng Li}{yifeng.li@nrc-cnrc.gc.ca}

\icmlkeywords{Anomaly detection, capsule net, normality score}

\vskip 0.3in
]



\printAffiliationsAndNotice{}  

\begin{abstract}
In this paper, we develop and explore deep anomaly detection techniques based on the capsule network (CapsNet) for image data. Being able to encoding intrinsic spatial relationship between parts and a whole, CapsNet has been applied as both a classifier and deep autoencoder. This inspires us to design a prediction-probability-based and a reconstruction-error-based normality score functions for evaluating the ``outlierness'' of unseen images. Our results on three datasets demonstrate that the prediction-probability-based method performs consistently well, while the reconstruction-error-based approach is relatively sensitive to the similarity between labeled and unlabeled images. Furthermore, both of the CapsNet-based methods outperform the principled benchmark methods in many cases.
\end{abstract}

\section{Introduction}
\label{introduction}

As real-time tracking \& diagnosis systems and autonomous controlling devices are strongly demanded in various domains in the current era of Internet of Things (IoTs), smart cities, big data, and deep learning, anomaly detection (also known as \textit{outlier detection}) is becoming increasingly critical. It aims at uncovering abnormal data points which may stand for novel or alarming events. Anomaly detection is of key importance in IoT systems, data centres, security platforms, and life science to diagnose system failure, detect intruders or attackers, and discover novel knowledge.   

Prior to the use of deep learning approaches, statistical and heuristic methods were main tools for anomaly detection in restricted application domains. Kernel methods, such as kernel density estimation (KDE) \cite{parzen1962,Kim2008} and support vector domain description (SVDD) \cite{Tax1999}, were the most successful ones to deal with non-linearity of the input feature space through the kernel trick.
However, nowadays the tremendous amount of data of various types (such as images, texts, omics data, etc.) have been collected, posing new challenges in handling the scalability and complexity of such data. With billions of samples in modern datasets, traditional methods become less effective. For example, conventional feature encoding and extraction methods are incompetent to capture informative factors from the input feature space of complex data. 

Embracing the wealth of data, deep learning models have achieved significant successes in various discriminative and generative modellings of modern data \cite{Bengio2003,Hinton2006b,Krizhevsky2012,Graves2013,Kingma2014,Goodfellow2014,Lecun2015}, encouraging to explore deep anomaly detection solutions. The core of deep learning is learning complex representations for the data at different levels in the latent space \cite{Bengio2013}. For example, convolutions on sequence data and embedding technologies on discrete data allow to encode visible examples to low-dimensional continuous dense vectors in a latent space, showing the advantage of distributed representation learning over alternative approaches.
Furthermore, stochastic gradient descent using mini-batches makes learning of deep networks very scalable to data of big size. The past few years have witnessed progresses made by the work reviewed in Appendix \ref{sec_related}. Accordingly, comparative studies, e.g. \cite{Skvara2018}, unconsidering these two advantages of deep learning over classic methods, could be biased.  


The technique in \cite{Golan2018} is based on the insightful observation: learning to discriminate between many types of geometrically transformed images encourages learning of features that are useful for detecting novelties. Among all geometric transformations, they only considered compositions of horizontal flipping, translations, and rotations, resulting in 72 distinct transformations. Their main focus was tackling the problem of identifying anomalous images in pure single class setting, even though it was mentioned their method may also be effective at distinguishing out-of-distribution samples from multiple-class data.

In fact, learning transformation is very challenging in computer vision tasks. Convolutional neural network (CNN) \cite{LeCun1998}, a hierarchy of convolution operations, has been widely used as a highly effective technique in classifying images. However, one arguable key limitation of CNN is that the neurons do not sufficiently capture the properties of entities such as position, orientation, and sizes, as well as their part-whole relationship. The capsule network (CapsNet)~\cite{Hinton2011,Sabour2017,Hinton2018} has been proposed and shown advantages in maintaining such information, which is a novel and promising structure that may be more closely related to biological neural organization. A capsule is a group of neurons whose activity vector represents the instantiation parameters of a specific type of entity such as an object or part. It has been demonstrated that CapsNet is capable of preserving hierarchical pose (position, size, and orientation) relationships between image features. For a given image, CapsNet can automatically and dynamically model affine transformations and part-whole relationships using an iterative routing-by-agreement mechanism. 

Inspired by these developments, we propose that, instead of using geometric transformations to learn distinct features of images, CapsNet can be employed to automatically learn transformations from the training examples such that a test example that cannot be explained by the network should be viewed as anomaly. Unlike \cite{Golan2018}, our work concentrates on the cases where normal samples come from multiple classes.
Our contributions are three-fold: (1) based on unique characteristics of CapsNet, we propose two normality score functions that work well; to the best of our knowledge, this is the first attempt to explore and test CapsNet for deep anomaly detection; (2) we provide insights and categorize existing ideas for deep anomaly detection into boundary-based and distribution-based families, paving the road for future studies; (3) we compared our methods with principled benchmark methods and assessed their capacities for deep anomaly detection. 

The rest of this paper is organized as follows. Insights into existing work, a gentle introduction to CapsNet, benchmarks, and data are provided in the Appendix \ref{sec_related}. The CapsNet-based normality score functions are described in Section \ref{sec_method}. In Section \ref{sec_exp}, the proposed methods are evaluated on three datasets in comparison with benchmarks. 

\section{CapsNet-Based Normality Score Functions} \label{sec_method}

Unlike the method in \cite{Golan2018}, we actually do not need to manually transform each training image. Using CapsNet, transformations ought to be automatically learned via an iterative routing-by-agreement mechanism. We can thus ignore the stage of labeling each transformed image. After a CapsNet classifier is trained, unseen images (either normal or out-of-distribution) could be directly fed into the learned model. We present two normality score function to determine the \textit{outlierness} of these images.

\begin{figure}[!htb]
    \centering
    \begin{subfigure}[t]{0.5\textwidth}
    	\centering
        \includegraphics[width=.95\textwidth]{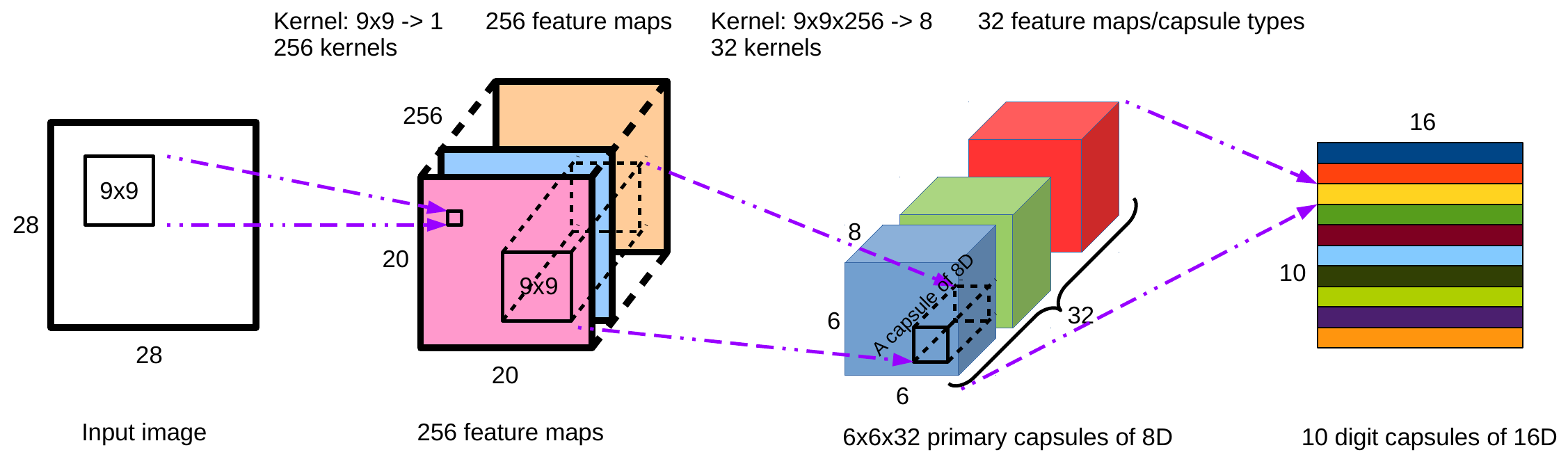}
        \caption{Encoder as classifier.}
        \label{fig_capsnetencoder}
    \end{subfigure}\\
    \begin{subfigure}[t]{0.5\textwidth}
    	\centering
        \includegraphics[width=.6\textwidth]{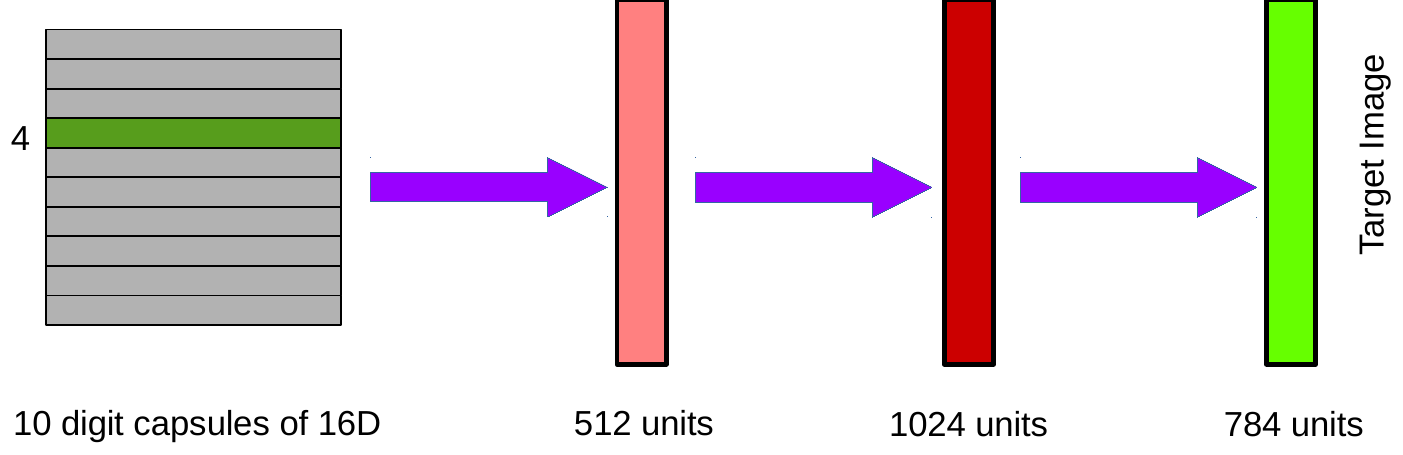}
        \caption{Decoder as regularizer.}
        \label{fig_capsnetdecoder}
    \end{subfigure}
    \caption{Architecture of the CapsNet designed in \cite{Sabour2017} for MNIST data. The main structure displayed in (\subref{fig_capsnetencoder}) is a classifier but can also be viewed as an encoder. The reconstruction regularizer can be viewed as a decoder as displayed in (\subref{fig_capsnetdecoder}).} \label{fig_capsnet}
\end{figure}

\subsection{Prediction-Probability-Based Normality Score}
The activation probabilities of digital capsules at the last layer of a CapsNet indicate the probabilities of the input sample belonging to the classes. However, unlike softmax probabilities in CNN, the activation probabilities of all digit capsules do not necessarily sum to one. Assuming the network is trained sufficiently well, for a normal test image, there should be one and only one probability being close to ``1'', representing the possibility of this image belonging to its true class. However, when an anomalous sample cannot be explained by the network, all activation probabilities of digital capsules would be very low. Therefore, this unique characteristic inspires us to define a normality score function $n_s(s)$ based on prediction probabilities (PP):

\begin{equation}
    n_{PP}(x) \triangleq \max\limits_{c=1, \cdots, C}(\lVert \bm{h}_c \lVert_2), 
    \label{eq_pp}
\end{equation}
where $\bm{x}$ is an input image, $\bm{h}_c$ represents the $c$-th digit capsule (see Appendix \ref{sec_related_capsnet}), $\lVert \bm{h}_c \lVert_2$ denotes the probability of $\bm{x}$ belonging to the $c$-th class. Hereafter, we simply call Equation (\ref{eq_pp}) \textit{PP score function}. 
Since the threshold, dividing the normal and the anomalous, is hard to decide, as per convention, we use the area under the receiver operating characteristic curve (auROC) to measure its performance.

\subsection{Reconstruction-Error-Based Normality Score}

In CapsNet, reconstruction error is used as a regularization term through a decoder component (Figure \ref{fig_capsnetdecoder}). The classifier is also an encoder (Figure \ref{fig_capsnetencoder}) for disentangled representation learning. In this perspective, a CapsNet thus meanwhile function as a deep autoencoder, offering an idea of score normality based on reconstruction error.

In this method, we use normalized squared error (NSE) to measure the quality of reconstructed images. The reason of using NSE instead of MSE (mean squared error) is that, different objects in different images (e.g. MNIST digits) can have different numbers of nonzero pixels in contrast with pure background. Using MSE will significantly weaken the actual difference between the input image and the reconstructed image. For example, as digit image ``1'' takes much less numbers of pixels than digit ``8'', using MSE the reconstruction loss of image ``1'' will be reduced at a greater extent than that of image  ``8'', even though image ``1'' may be reconstructed worse than image ``8''. The reconstruction error (RE) based normality score can thus be defined as:
\begin{equation}
    n_{RE}(\bm{x}) \triangleq - \text{NSE}(\bm{x}') = - \frac{\lVert \bm{x} - \bm{x}' \rVert_2^2}{\lVert \bm{x} \Vert_2}
\end{equation}
where $\bm{x}$ represents an actual image and $\bm{x}'$ the reconstructed image. When the background in an image takes a large portion of the space, and the values of background pixels are near zeros, the advantage of NSE will be more obvious. Hereafter, we refer to Equation (\ref{eq_pp}) as \textit{RE score function}.

\section{Experiments} \label{sec_exp}

\subsection{Case Studies in Normality Score Functions}
We looked into the performance of our two normality score functions through two case studies on the MNIST data.

\subsubsection{Multi-Class Training Data and Single-Class Anomalous Digits} 
\begin{figure}[!htb]
    \centering
    \begin{subfigure}[t]{0.49\linewidth}
    	\centering
        \includegraphics[width=\linewidth]{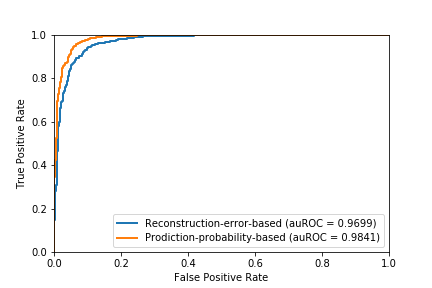}
        \caption{Digit 2 as anomalous class.}
        \label{fig_anomalous2}
    \end{subfigure}
    \begin{subfigure}[t]{0.49\linewidth}
    	\centering
        \includegraphics[width=\linewidth]{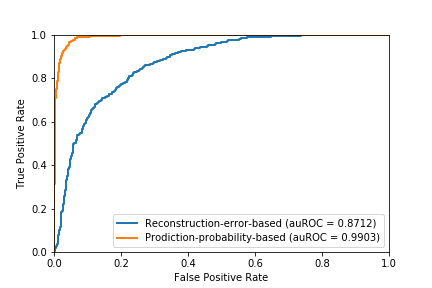}
        \caption{Digit 9 as anomalous class.}
        \label{fig_anomalous9}
    \end{subfigure}
    \caption{ROC curves of detecting anomalous digits: ``2'' (\subref{fig_anomalous2}) and ``9'' (\subref{fig_anomalous9}) respectively using CapsNet.} \label{fig_anomalous29}
\end{figure}

\begin{figure}[!htb]
    \centering
    \begin{subfigure}[t]{0.49\linewidth}
    	\centering
        \includegraphics[width=.8\linewidth]{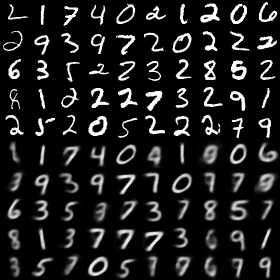}
        \caption{Digit 2 as anomalous class.}
        \label{fig_digit2}
    \end{subfigure}
    \begin{subfigure}[t]{0.49\linewidth}
    	\centering
        \includegraphics[width=.8\linewidth]{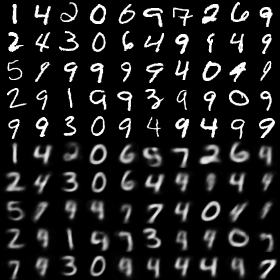}
        \caption{Digit 9 as anomalous class.}
        \label{fig_digit9}
    \end{subfigure}
    \caption{Original digits (upper half) and reconstructed digits (lower half), when detecting anomalous digits ``2'' (\subref{fig_digit2}) and ``9'' (\subref{fig_digit9}) respectively using CapsNet.} \label{fig_digit29}
\end{figure}

Figure \ref{fig_anomalous29} shows the performance of our methods when selecting digits ``2'' and ``9'' as abnormal samples, respectively, and the rest digits as normal samples. When digit ``2'' was treated as anomaly class, both PP and RE score functions achieved near perfect auROCs (0.9841 and 0.9699).  When ``9'' was viewed as anomaly class, however, the PP score function outperformed the RE score function by auROC of 0.12. The reason for this can be found from Figure \ref{fig_digit29}, which depicts $50$ real digits and their reconstructed ones for both digits. From Figure \ref{fig_digit29}, one can easily notice that anomalous digit ``9'' was mostly reconstructed as digit ``4'', while anomalous digit ``2'' was reconstructed as several different digits such as ``1'', ``3'', ``6'', and ``7''. As images ``4'' and images ``9'' are quite similar, the normality scores of images ``9'' become very high, even though the true digits ``9'' and their reconstructed versions ``4'' are two different numbers. 

\subsubsection{Multi-Class Traing Data and Multiple Anomalous Digits} 
\begin{figure}[!htb]
       \centering
        \includegraphics[width=.5\linewidth]{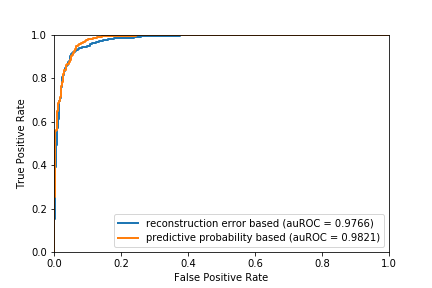} 
        \caption{ROC curves of detecting abnormal 
        digits ``0'', ``3'', and ``5'' using our two CapsNet-based score functions.}
         \label{ROC035}
    \end{figure}

In this case, we tested our two CapsNet-based normality score functions by considering digits ``0'', ``3'' and ``5'' as abnormal digits, and the rest as normal. The results is displayed in Figure \ref{ROC035}. Both methods achieved similar auROCs.

\subsection{Comparison on MNIST Dataset}
\begin{figure}[!htb]
       \centering
        \includegraphics[width=0.8\linewidth]{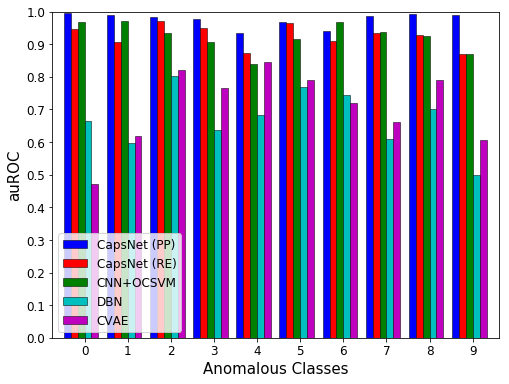} 
         \caption{Performance on the MNIST data.}
         \label{fig_auROC_MNIST}
    \end{figure}

Our two methods and three benchmarks CNN+OCSVM, DBN, and VAE (see Appendix \ref{sec_benchmarks}) were compared on MNIST data by respectively treating each digit class as anomalous class and the rest as normal. Their performance in terms of auROC is displayed in Figure \ref{fig_auROC_MNIST}. One can see that our CapsNet-based PP method works consistently the best, while our CapsNet-based RE method in generally is slightly inferior to the PP method, but has comparable results as CNN+OCSVM. The two DGMs (DBN and VAE) are not competitive to the CapsNet-based and deep hybrid methods.  

\subsubsection{Comparison on Fashion-MNIST Dataset}
When comparing all five methods on the Fashion-MNIST data, generally speaking all methods tended to get lower results, which is reasonable because Fashion-MNIST samples is more complicated than MNIST samples. When using footwear (Sandals, Sneakers, or Ankle boots) as anomalous samples, the CapsNet(RE) method outperformed the CapsNet(PP) method. It is because the CapsNet classifier could carry some wrong confident information of classifying a footwear anomalous sample (e.g. Sneaker) to the normal footwear classes (e.g. Sandal or Ankle boots), while reconstruction errors can pick up differences in details. In the case of using data from a topwear class as anomalous samples, CapsNet(PP) worked better than CapsNet(RE). When trousers and bags were viewed as abnormal samples, both methods worked quite well without big differences in performance. The performances of CNN+OCSVM and VAE vary largely. In few cases, they can obtain similar results as CapsNet-based methods. DBN did not behave impressively on the data.

     \begin{figure}[!htb]
       \centering
        \includegraphics[width=0.8\linewidth]{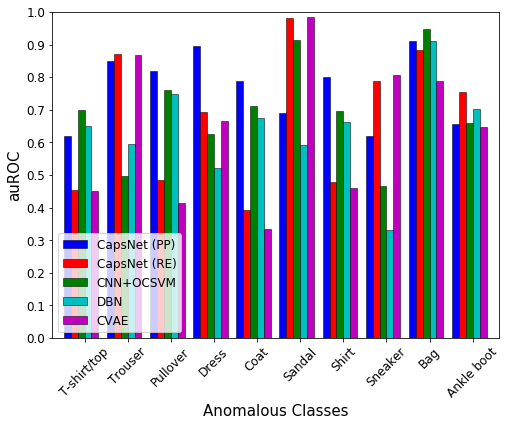} 
         \caption{Performance on the Fashion-MNIST data.}
         \label{fig_auROC_FashionMNIST}
    \end{figure}
    
\subsection{Comparison on Small-Norb Dataset}
Small-Norb is a challenging data for all five methods. When using animals and cars as anomalies, only CapsNet(PP) performed reasonably good, the other methods behaved randomly. When trucks were used as anomalous samples, only CapsNet(PP) and CapsNet(RE) were able to behave non-randomly. In the case of planes as abnormal data, both CapsNet(RE) and VAE worked the best. Only in the case of humans as anomalies, CNN-OCSVM reached 0.7 auROC.

     \begin{figure}[!htb]
       \centering
        \includegraphics[width=0.8\linewidth]{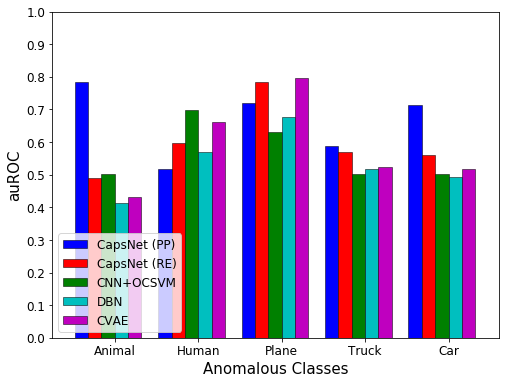} 
         \caption{Performance on the Small-Norb data.}
         \label{fig_auROC_Small-Norb}
    \end{figure}

\section{Conclusions and Future Work} \label{sec_con}

Many modern data-driven intelligent systems require more accurate anomaly detection techniques. In this paper, we explore novel solutions in consideration of CapsNet's distinct characteristics. We devise two normality score functions based on CapsNet's activation probability and reconstruction error respectively. Experiments on three image data sets show that both methods have complementary strengths and outperform existing solutions in many setups.


In this paper, we did not discuss classification-based anomaly detection methods, as they essentially treat an anomaly detection task as a two-category classification problem by using both normal and abnormal samples in the training process. While all out-of-distribution anomaly detection methods as discussed in this paper only take normal samples for training, because an anomalous data point could unpredictably come from anywhere outside the normal data distribution, which is true in many application domains. 
We did not discuss deep reinforcement learning methods for anomaly detection in this paper as well. But this is a new area worthy of future investigation.
The performance of our normality score functions depends on CapsNet's learning capacity on certain data. However, as mentioned in \cite{Sabour2017}, same as deep generative models, current CapsNets do not perform very well when the backgrounds of images vary too much (such as CIFAR-10) to be modelled. Hence combining our two normality score functions with other solutions mentioned in Appendix \ref{sec_related} may lead to a more robust solution.

\bibliography{myRef.bib}
\bibliographystyle{icml2019}

\appendix
\section{Insights into Existing Work} \label{sec_related}
Instead of simply enumerating all existing work for deep anomaly detection, we categorize these solutions into two families, and provide insights into their characteristics, strengths and challenges.

A classical viewpoint regarding anomaly detection is that, learning the boundary of data mass is more effective and straightforward than learning the density distribution of the data, because (1) available data were too few to cover the distribution in many cases, and (2) it was much more complicated and difficult to model data distribution using a generative model. However, in the big data era, massive amount of data become available in many domains; many of the data are structured (such as images, graphs, time-series, text, etc.); and deep generative models have achieved promising successes in modelling such modern data, offering a new avenue for exploring distribution-based methods for anomaly detection. These two categories, boundary-based and distribution-based, are respectively discussed below. 

\subsection{Boundary-Based Methods}
Kernel based support vector domain description (SVDD) or one-class support vector machine (OCSVM) methods, including hypersphere \cite{Tax1999} and hyperplane \cite{Scholkopf2001a} models, is a successful family for anomaly detection in the pre-deep-learning era. The idea of the hypersphere-based SVDD is to map data points from input space to high-dimensional space and learn a hypersphere that capture the core mass of the data distribution. Any data point outside the hypersphere is viewed as an abnormal sample. Please see \cite{LiThesis2013,LiNeuro2015} for a systematic discussion of SVDD methods. It is quite natural to think of deep extension of these methods to continue their success in the deep learning age. In pursuit of this aim, there are two efforts: deep hybrid methods and one-class neural network models. In deep hybrid methods (e.g. VAE+OCSVM \cite{Andrews2016} and DBN+OCSVM \cite{Erfani2016}), a supervised (e.g. CNN or recurrent net) or unsupervised (e.g. deep belief net (DBN) or varitiaonal autoencoder (VAE)) neural network is first employed to learn embedding representations of samples in the hidden space, then using these latent representations as inputs an SVDD method is used to detect abnormal data points.  The one-class neural network models (e.g. one-class deep SVDD \cite{Ruff2018}) learn a neural network and SVDD together by maximising an adapted SVDD objective in the prime form. Both classes of methods have pros and cons. Deep hybrid methods are pipelines that are very flexible in choosing and combining different representation learning (or pretrained embeddings) and SVDD models. However, these methods face the challenge of scalability due to size of kernel matrices in dual form of SVDD. Deep SVDD models explicitly use deep neural networks as feature extractors in replacement of implicit kernel tricks, and are scalable to large data due to use of SVDD's prime forms and stochastic gradient descent. Nevertheless, this strategy is lack of flexibility in practice, that is a specific model needs to be built for each specific type of data.

\subsection{Distribution-Based Methods}
As mentioned above, deep generative models (DGMs), such as deep belief net (DBN) \cite{Hinton2006a,LiICML2018} and variational autoencoder (VAE) \cite{Kingma2014}, can be applied as unsupervised feature learning techniques. More importantly, since DGMs aim at modelling the joint distribution of visible and latent variables (that is $p(\bm{x},\bm{h})$), their likelihood $p(\bm{x})$ by marginalising out $\bm{h}$ may serve as an anomality describer. However, exact likelihood can only be obtained in quite few generative models, such as exponential family restricted Boltzmann machines (RBMs) \cite{LiIJCNN2018}. In many cases when a recognition component is used for approximate inference, only the variational/evidence lower bound (ELBO) of log-likelihood is available. Unfortunately, ELBO may be too loose to be an normality indicator. Oftentimes, the ELBOs of normal and abnormal samples indistinguishably fall into the same range.  Luckily, this is not the end of the story. In an architecture with encoder (recognition) component and decoder (generative) component, reconstruction error could serve as anomality measure based on the intuition that out-of-distribution samples can be reconstructed badly \cite{An2015}. But in the case of generative adversarial net (GAN) based methods \cite{Schlegl2017}, an encoder is unavailable. For an inquiry sample $\bm{x}$, a supervised learning process has to be executed to search for a hidden representation $\bm{h}'$ such that the corresponding generated sample $\bm{x}'$ best approximates $\bm{x}$. The approximation error and probability from GAN's discriminator can together indicate the extend of anomality.

\subsection{Capsule Nets}
\label{sec_related_capsnet}
The concept of capsule was first introduced in transforming autoencoder for image modelling \cite{Hinton2011}. But its potential capacity was not demonstrated until the publications of vector capsule net (CapsNet) \cite{Sabour2017} and matrix capsule net \cite{Hinton2018}. Since both models are conceptually very similarly, we used vector CapsNet in our studies. The normality score functions described in Section \ref{sec_method} should also work for matrix CapsNet.  Figure \ref{fig_capsnet} shows an example of such network for the MNIST data where each capsule is a vector of 8 neurons and a digit capsule has 16 neurons. Each capsule is a sparse linear combination of all transformed vectors of lower capsules. A capsule can encapsulate pose information (such as position, orientation, scaling, and skewness) and instantiation parameters (such as color and texture). The connections and activations between a parent capsule and child capsules are dynamically determined, thus transformations and part-whole relationships can be modeled, implying a revolutionary potential for structured data modeling (such as images, videos, and documents). 

Formally, the $j$-th capsule at layer $l+1$ can be computed as
\begin{align}
\bm{h}_j^{(l+1)} = \frac{\lVert \bm{s}_j^{(l+1)} \rVert_2^2}{1+\lVert \bm{s}_j^{(l+1)} \rVert_2^2} \frac{\bm{s}_j^{(l+1)}}{\lVert \bm{s}_j^{(l+1)} \rVert_2},
\label{eq_squash}
\end{align}
where $\bm{s}_j^{(l+1)}$ is a sparse linear combination of pose vectors from the lower layer:
\begin{align}
\bm{s}_j^{(l+1)} = \sum_{i=1}^{K_l} c_{i,j} \bm{h}_{i,j}^{(l)},
\end{align}
where $K_l$ is total number capsules at level $l$, and a pose vector $\bm{h}_{i,j}^{(l)}$ is transformed from a capsule:
\begin{align}
\bm{h}_{i,j}^{(l)} = \bm{W}_{i,j}\bm{h}_i^{(l)} \quad \forall i \text{ in } \{ 1,\cdots,K_l \},
\end{align}
where $\bm{W}_{i,j}$ is a transformation matrix to be learned using back-propagation, $\bm{h}_i^{(l)}$ is the $i$-th capsule at level $l$, and $c_{i,j}$ is called a coupling coefficient and is computed using a softmax function
\begin{align}
c_{i,j} = \frac{e^{b_{i,j}}}{\sum_{j'=1}^{K_{l+1}} e^{b_{ij'}} },
\end{align}
where $b_{i,j}$ is determined using a dynamic routing algorithm based on inner product (cosine) ${\bm{h}_j^{(l+1)}}^T \bm{h}_{i,j}^{(l)}$. 

The squash function in Equation \ref{eq_squash} allows the length of a capsule, $\lVert \bm{h}_j^{(l+1)} \rVert_2$, to serve as its activation probability. Thus, there is no explicit activation unit in vector CapsNet.

The objective function is defined by a margin loss which uses $l_2$ norms of digit capsules as prediction probabilities:
\begin{align}
L = \sum_{c=1}^C L_c,
\end{align}
where $C$ is the total number of classes, and
\begin{align}
&L_c = \delta(y - c) \max(0, m_+ - \lVert \bm{h}_c \rVert_2)^2\nonumber\\
&+ \lambda (1-\delta(y - c)) \max(0, \lVert \bm{h}_c \rVert_2 - m_-)^2,
\end{align}
where $y$ is actual class label, $\bm{h}_c$ is a digit capsule, $\delta(y - c) = 1$ if and only if $y=c$, $m_+=0.9$, $m_-=0.1$, and $\lambda=0.5$.  

Dynamic routing is used between the primary capsule layer and the digital capsule layer. Thus, in Figure \ref{fig_capsnet}, the $6\times 6 \times 32$  lower-level capsules dynamically connect to the 10 higher-level digit capsules. Since bottom-up activation paths form a tree-like structure and a very deep tree is impractical, CapsNet should not have many capsule layers.

The architecture of CapsNet used in this paper is identical to that of \cite{Sabour2017}. The number of epochs was set to $20$ for experiments on MNIST dataset, and $50$ for both Fashion-MNIST and Small-Norb datasets; batch size was chose as $100$ for MNIST and Fashion-MNIST datasets and $64$ for Small-Norb dataset.

\section{Benchmarks}
\label{sec_benchmarks}
Our CapsNet based anomaly detection methods were compared with three principled methods from the two family of methods discussed in Section \ref{sec_related}. These benchmarks are described as bellows. Please note that the method presented in \cite{Golan2018} is unavailable for multi-class normal data, thus we were unable to compare with it.

\begin{itemize}
    \item We implemented a deep hybrid method named CNN+OCSVM, obviously CNN was used to learn the latent representations and OCSVM was used for anomaly detection. The CNN component has two convolutional layers (3$\times$3 receptive fields in both layers, 32 and 64 feature maps respectively for the first and second layers, max-pooling with pooling size of 2$\times$2 after the second layer, ReLU activation function), one fully connected layer (128 units), and a softmax layer for class labels of normal samples. The outputs of the fully connected layer were latent representations extracted for searching the hyperparameters and optimizing the model parameters of OCSVM.
    \item We employed a three-layer DBN to model data distribution and then measured reconstruction error to score abnormality. There are 500 units in each hidden layer. The model was layer-wise pretrained by RBMs. Bernoulli distribution was assumed for both visible and hidden layers. The pixel values were scaled to interval [0,1] as input to DBN. Reconstruction error of an inquiry sample was used to detect anomalous samples.
    \item Similarly, a convolutional VAE was also applied to capture data distribution.  The inference component (encoder) has the same architecture as the CNN component (disregarding the output layer) in CNN+OCSVM. The latent space size was set to 64. The structure of the inference component was mirrored for the structure of the generative component. Reconstruction error was used in determination of anomality.
\end{itemize} 

\section{Datasets}
\label{sec_data}
Our two methods and the three benchmarks were evaluated on three image datasets with increasing difficulty: MNIST, Fashion-MNIST and Small-Norb. In learning stages, all methods were trained using same normal training examples. In test processes, same normal samples and anomalous data were used. We kept the normal and anomalous test data balanced, so that we can focus on comparison using auROC and leave imbalanced issue for future investigation.
\begin{itemize}
    \item \textbf{MNIST}: It contains a training set of $60000$ $28 \times 28$ grayscale digit images and a test set of $10000$ same resolution grayscale examples from approximately $500$ different writers \cite{LeCun1998}.
    \item \textbf{Fashion-MNIST}: It is a dataset of Zalando's article images, comprising $70000$ $28 \times 28$ MNIST-like labeled fashion images with $7000$ images per category  \cite{Xiao2017}. The training set has $60000$ images and the test set has $10000$ images. The samples come from $10$ classes: T-shirt/top, trouser, pullover, dress, coat, sandal, shirt, sneaker, bag, and ankle boot.
    \item \textbf{Small-Norb}: It contains $24300$ $96 \times 96$ grayscale images pairs of $50$ toys belonging to $5$ generic categories: four-legged animals, human figures, airplanes, trucks, and cars \cite{LeCun2004}. The objects were imaged by two cameras under $6$ lighting conditions, $9$ elevations , and $18$ azimuths. As in \cite{Sabour2017}, the images were resized to 48$\times$48; random 32$\times$32 crops of them were obtained during training process. Central 32$\times$32 patches of test images were used during test.
\end{itemize}

\end{document}